\documentclass[runningheads]{llncs}

\usepackage{eccv}

\usepackage{multirow}

\usepackage{eccvabbrv}

\usepackage{graphicx}
\usepackage{booktabs}

\usepackage[accsupp]{axessibility}  

\usepackage{hyperref}

\usepackage{orcidlink}

\begin{document}

\title{Indoor 3D Reconstruction with an Unknown Camera-Projector Pair} 

\titlerunning{Indoor 3D reconstruction with a unknown CPP}

\author{Zhaoshuai Qi \and
Yifeng Hao \and Rui Hu \and Wenyou Chang \and Jiaqi Yang\inst{*} \and Yanning Zhang\inst{*}}

\authorrunning{Z. Qi, Y. Hao, W. Chang et al.}

\institute{
National Engineering Laboratory for Integrated Aero-Space-Ground-Ocean Big Data Application Technology and the College of Computer Science, Northwestern Polytechnical University, Xi’an, ShaanXi, 710072 China\\
\email{zhaoshuaiqi1206@163.com}\\
\email{\{jqyang,ynzhang\}@nwpu.edu.cn}}

\maketitle

\begin{abstract}
 Structured light-based method with a camera-projector pair (CPP) plays a vital role in indoor 3D reconstruction, especially for scenes with weak textures. Previous methods usually assume known intrinsics, which are pre-calibrated from known objects, or self-calibrated from multi-view observations. It is still challenging to reliably recover CPP intrinsics from only two views without any known objects. In this paper, we provide a simple yet reliable solution. We demonstrate that, for the first time, sufficient constraints on CPP intrinsics can be derived from an unknown cuboid corner (C2), e.g. a room’s corner, which is a common structure in indoor scenes. In addition, with only known camera principal point, the complex multi-variable estimation of all CPP intrinsics can be simplified to a simple univariable optimization problem, leading to reliable calibration and thus direct 3D reconstruction with unknown CPP. Extensive results have demonstrated the superiority of the proposed method over both traditional and learning-based counterparts. Furthermore, the proposed method also demonstrates impressive potential to solve similar tasks without active lighting, such as sparse-view structure from motion.
\end{abstract}

\section{Introduction}
\label{sec:intro}
Compared with other 3D reconstruction methods, such as structure from motion (SfM) \cite{r2}, multi-view stereo (MVS) \cite{r1,r3,r4}, time-of-flight (TOF) cameras \cite{ri}, structured-light (SL) methods \cite{r16} can produce accurate and dense 3D point cloud of the scene without requiring texture on the scene surfaces. In addition, the system used in SL is simple and cheap, which often consists of a camera-projector pair (CPP). These advantages make SL more suitable for indoor scenes, even when dealing with walls or floors that lack textures. However, most CPPs are assumed pre-calibrated offline \cite{r19} before 3D reconstruction. The parameters of the camera and projector are known and fixed across reconstruction. This limits its flexibility and applications, especially for scenarios where parameters of the CPP need to be frequently adjusted. It is more desirable to reconstruct UNKNOWN indoor scenes with an UNKNOWN CPP, where the underlying CPP self-calibration is still an open problem due to the following challenges.

\begin{figure}[htb]
    \centering
    \includegraphics[width=0.85\linewidth]{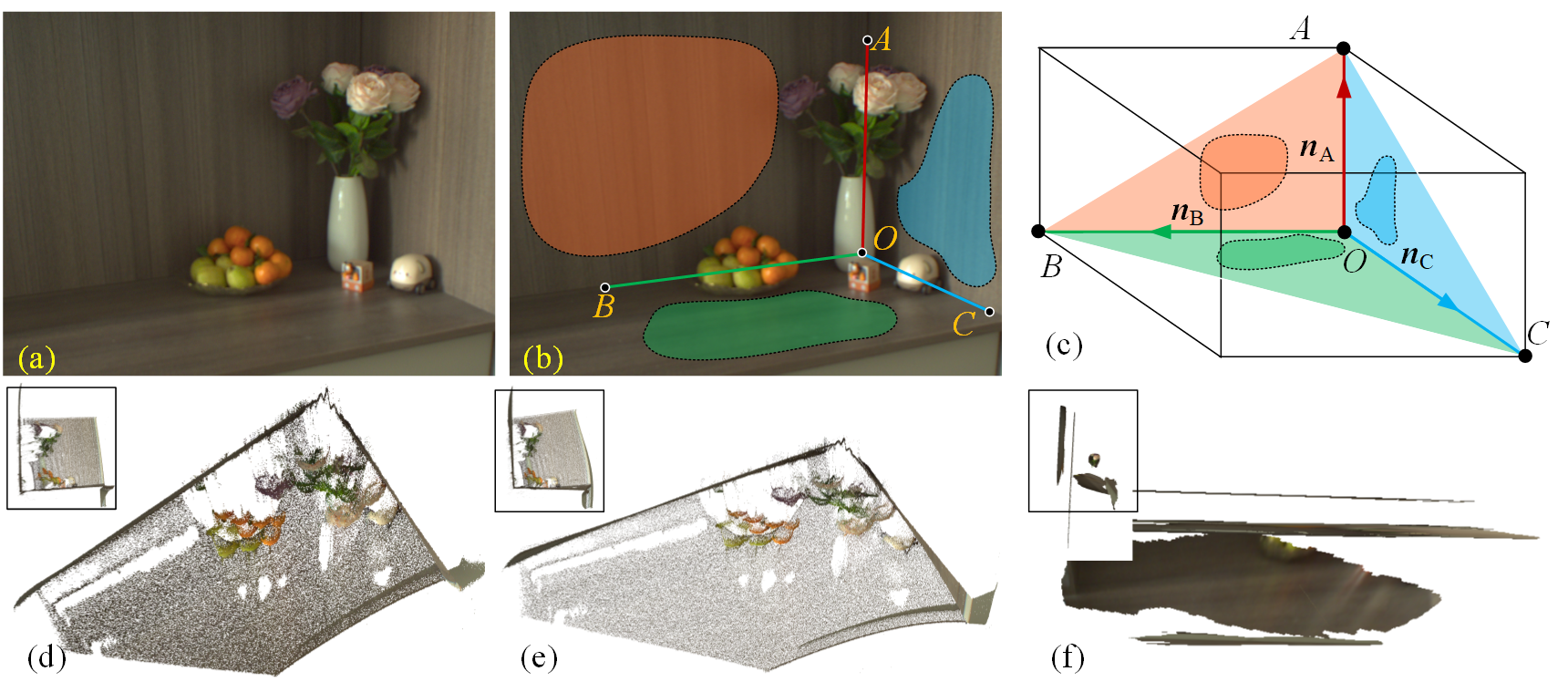}
    \caption{Reconstructing indoor scenes with an unknown CPP is challenging, since the self-calibration only from two views is ill-posed and that few scene cues are available due to texture-lessness. We propose to leverage an unknown cuboid corner (C2) to extract sufficient constraints for calibration. As shown in \cref{fig1}(a), a C2 can be a room corner with partially observed walls and floor as its faces, see \cref{fig1}(b), or, more formally, a Tri-rectangular tetrahedron in \cref{fig1}(c). Compared with two-view COLMAP \cite{r2} and PlaneFormer \cite{r12}, which struggle to reconstruct the scene with significant distortion, see \cref{fig1}(d) and \cref{fig1}(e), our method achieves impressive result, see \cref{fig1}(f).}
    \label{fig1}
\end{figure}

Firstly, CPP self-calibration can be formulated as a typical two-view vary-intrinsic camera self-calibration problem, where the projector in a CPP is treated as another camera with distinct parameters. This problem is under-constrained for previous camera self-calibration in restricted\cite{rc1,rc2} or general motion\cite{rc3}, since only two constraints can be constructed from the Kruppa equation \cite{rc5,rc6,rc7} or fundamental matrix \cite{rc8,rc9}, whereas there are at least three unknowns even for a constant camera. Therefore, at least three views \cite{rc4} (often tens or more for a reliable result) are required for previous methods including most learning-based \cite{rc19,rc20,rc21,rc22}. In addition, the assumptions, such as a constant camera, or a known principal point (PP) \cite{rc10,rc11,rc12,rc13}, do not hold for a CPP, where the intrinsics of a projector are often different from those of the camera. The unknown PP and focal length for the projector plus unknowns of the camera (even with a known PP) still make the calibration an challenging ill-posed problem. 

Secondly, there are only limited scene cues for CPP self-calibration in indoor scenes. Indeed, for an indoor scene, there are lots of planes, from which inter-view homographies \cite{rc15,rc16} provide additional constraints. However, at least four such homographies yield a complete solution, which is incapable of the two-view CPP problem. The Manhattan World (MW) assumption \cite{r7} is another common scene property. Under this assumption, vanishing points (VPs) can be extracted from images of parallel line segments \cite{rc18}. At least three (say mutually orthogonal) VPs yield a camera calibration \cite{rc17}, even from a single image. Nevertheless, the texture-lessness of an indoor scene, see \cref{fig1}(a), often “violates” this assumption, leading to insufficient lines for reliably estimating a VP.

Please note that recent learning-based methods \cite{r8,r9,r10,r11,r12,r13,r14} for indoor reconstruction from sparse or even single view have demonstrated impressive performance. However, most of them still assume a known or partially (PP is known) known camera. 

We propose a method to overcome these limitations above. Instead of line segments, we focus on another common but simple structure in indoor scenes, i.e. a cuboid corner (C2) such as a room corner (see \cref{fig1}(a)), from which sufficient constraints can be constructed for CPP self-calibration. The geometry of a C2 needs not be known, and occlusion is allowed when observing it. Additionally, with only known camera principal point, the complex multi-variable estimation of all CPP intrinsics can be simplified to a simple univariable optimization problem. This leads to a reliable and accurate calibration from only two-view observations of an unknown scene, see \cref{fig1}(d), achieving indoor 3D reconstruction with only an unknown CPP. Furthermore, this C2-based calibration can be easily extended to similar tasks, such as the challenging camera self-calibration in two-view SfM.

\section{Notations and Basic Principles}
\subsection{Camera model}

A camera is represented with a pin-hole model as 
\begin{equation} 
\lambda\begin{bmatrix}
  x \\
  y \\
  1
\end{bmatrix}=\textbf{M}\begin{bmatrix}
  X \\
  Y \\
  Z \\
  1
\end{bmatrix},
\label{eqa1}
\end{equation}

\noindent where ${\mathbf M}$ is the camera matrix and $\lambda$ is a scalar.

The camera matrix is decomposed into a concatenation of camera intrinsics ${\mathbf K}$ and extrinsics, i.e. camera rotation ${\mathbf R}$ and translation $ \boldsymbol{t}$, as
\begin{equation} 
\textbf{M}=\textbf{K}\begin{bmatrix}
 \textbf{R} &&|&&\boldsymbol t
\end{bmatrix},
\label{eqa2}
\end{equation}

\noindent where ${\mathbf K}$ has the form as 
\begin{equation} 
\textbf{K}=\begin{bmatrix}
  f_{\rm x} & \gamma & x_{\rm 0} \\
  0 & f_{\rm y} & y_{\rm 0} \\
  0 & 0 & 1
\end{bmatrix},
\label{eqa3}
\end{equation}
\noindent where $f_{\rm x}$ and $f_{\rm y}$ are focal lengths along $x$ and $y$ directions. $\gamma$ is the skew factor. $\boldsymbol{x}_{0}=(x_{0},y_{0})$ is the principal point.

Similar to \cite{rc4}, we assume a “natural” camera with square pixels, i.e. $\gamma$ equals to 1 and $f_{\rm x} =f_{\rm y}=f$. Additionally, the lens distortion is assumed insignificant or corrected in advance, which is thus not considered in the following discussion.

\subsection{C2 parameterization}
\label{sec. C2para}
As shown in \cref{fig1}(c), a typical C2 is a Tri-rectangular tetrahedron with four vertices, e.g. $O, A, B, C$. Each two of vertices defines an edge and hence a total of six edges. There is a special vertex, i.e. $O$ in \cref{fig1}(c), usually called the right angle (RA), where the angle between each two edges with RA as their common vertex is a right angle. Accordingly, we have three mutually orthogonal edges for a C2, as shown in \cref{fig1}(c), which are called legs $\boldsymbol{n}_{\rm A}$, $\boldsymbol{n}_{\rm B}$ and $\boldsymbol{n}_{\rm C}$, respectively. The planes defined by each two of these legs are also orthogonal to each other, which are denoted as the faces $\rm{\Pi }_{\rm A}, \rm{\Pi } _{\rm B}$ and $\rm{\Pi } _{\rm C}$ with normalized $\boldsymbol{n}_{\rm A}$, $\boldsymbol{n}_{\rm B}$ and $\boldsymbol{n}_{\rm C}$ as their normal respectively, see triangles $\triangle ABO$ and $\triangle ACO$ in \cref{fig1}(c). Notably, significant occlusion of a C2 is allowed, as shown in \cref{fig1}(b), where all or some of the vertices and legs are not directly observed. In fact, we require only three partially observed faces of a C2 for calibration, since vertices and legs can be inferred from inter-plane homographies (see \cref{subsec.impl}). 

According to the definition above, a C2 is defined by seven parameters up to an unknown scale: three for rotation, two for scaled translation and two for the length ratios between legs. Specifically, in the camera coordinate frame, these parameters can be specified as 
\begin{equation}
\textbf{R}=\begin{bmatrix} 
 \frac{\boldsymbol{n}_{\rm A}}{\left \| \boldsymbol{n}_{\rm A}  \right \| } 
 & \frac{\boldsymbol{n}_{\rm B}}{\left \| \boldsymbol{n}_{\rm B}  \right \| } 
 &\frac{\boldsymbol{n}_{\rm C}}{\left \| \boldsymbol{n}_{\rm C}  \right \| } 
\end{bmatrix},
\label{eqa4}
\end{equation}

\begin{equation}
\boldsymbol{t}=\lambda_{\rm O} \boldsymbol{X}_{\rm O},
\label{eqa5}
\end{equation}
and 
\begin{equation}
\left \| \boldsymbol{n}_{\rm C}  \right \|:
\left \| \boldsymbol{n}_{\rm B}  \right \|:
\left \| \boldsymbol{n}_{\rm A}  \right \|
=k_{\rm C}:k_{\rm B}:1,
\label{eqa6}
\end{equation}

\noindent where $\boldsymbol{X}_{\rm O}$ is the 3D coordinate of $ O $, and $\lambda_{\rm O}$ is an arbitrary scalar. $k_{\rm B}$ and $k_{\rm C}$ are length ratios of corresponding legs, respectively.


\subsection{Problem formulation}

The core of direct 3D reconstruction with an unknown CPP lies in the CPP self-calibration. Provided a known camera principal point, the CPP self-calibration aims to recover the remaining CPP intrinsics from only two-view correspondences of an unknown C2.

More specifically, given matched image point pairs $\{\boldsymbol{x}_{\mathrm{c,S}}(i),\boldsymbol{x}_{\mathrm{p,S}}(i)\mid i{=}1,2,\\ \ldots,N_{\mathrm{S}}\}$ on faces $\rm{\Pi } _{\rm S}$ of a C2, $ S= A, B$ or $C$, respectively, and the camera principal point $\boldsymbol{x}_{\mathrm{c,0}}$, we aim to recover both focal lengths $f_{\rm c}$ and $f_{\rm p} $ of the camera and projector, and the principal point $\boldsymbol{x}_{\mathrm{p,0}}$ of the projector. Extrinsics and the parameters of C2 are last considered in this paper, since they can be easily determined after intrinsics estimated.

Please note that, even $\boldsymbol{x}_{\mathrm{c,0}}$ is assumed known, according to a simple argument counting, the self-calibration is still significantly ill-posed: There are four unknowns, i.e. $f_{\rm c}$, $f_{\rm p}$ and $\boldsymbol{x}_{\mathrm{p,0}}$, whereas the Kruppa equation can only provides two constraints, leaving two unknowns unconstrainted. For textureless indoor scenes with insufficient line segments, we will show how to solve this problem by constructing sufficient constraints from an unknown C2 in the following section.

\section{Geometry of cameras viewing a C2}
\label{sec:Princ}
In this section, we describe the single- and two-view geometry of a camera viewing a C2, from which the self-calibration algorithm is developed.
\subsection{Single-view geometry}
For clarity, we assume a canonical C2 as shown in \cref{fig1}(b). Suppose the four vertices $O$, $A$, $B$, $C$ of this C2 are imaged by a camera at image points $\boldsymbol{x}_\mathrm{O}$, $\boldsymbol{x}_\mathrm{A}$, $\boldsymbol{x}_\mathrm{B}$ and $\boldsymbol{x}_\mathrm{C}$, respectively. $\boldsymbol{X}_\mathrm{O}$, $\boldsymbol{X}_\mathrm{A}$, $\boldsymbol{X}_\mathrm{B}$ and $\boldsymbol{X}_\mathrm{C}$ are inhomogeneous 3D coordinates of these vertices, respectively, in the camera coordinate frame. The camera intrinsic matrix is $\mathbf{K}$. Please note that since only a single view is considered in this subsection, the subscript “c” is omitted.

Each vertex can be obtained by back-projecting from its image as 
\begin{equation}
\boldsymbol{X}_{\rm S}=\lambda_{\rm S} \textbf{K}^{ -1} \boldsymbol{x}_{\rm S}
\label{eqa7}
\end{equation}

\noindent where $S$ is the vertex index which can be $ O $, $ A $, $ B $ or $ C $.  $\lambda_{\rm S}$ is an unknown scalar.

Since the C2 is parameterized up to an unknown scalar, without loss of generalization, we set $ \lambda_{\mathrm{O}}$ to a specific value, say $\lambda_{\mathrm{O}}=1$. Accordingly, only $\lambda_{\mathrm{A}}$, $\lambda_{\mathrm{B}}$ and $\lambda_{\mathrm{C}}$ are unknown. 

By enforcing the orthogonality constraints between the legs $\boldsymbol{n}_{\rm A}$, $\boldsymbol{n}_{\rm B}$ and $\boldsymbol{n}_{\rm C}$, we have the following equations.
\begin{equation}
\begin{cases}\boldsymbol{n}_\mathrm{A}^\mathrm{T}\bullet \boldsymbol{n}_\mathrm{B}=\left(\boldsymbol{X}_\mathrm{A}-\boldsymbol{X}_\mathrm{O}\right)^\mathrm{T}\left(\boldsymbol{X}_\mathrm{B}-\boldsymbol{X}_\mathrm{O}\right)=0\\\boldsymbol{n}_\mathrm{B}^\mathrm{T}\bullet \boldsymbol{n}_\mathrm{C}=\left(\boldsymbol{X}_\mathrm{B}-\boldsymbol{X}_\mathrm{O}\right)^\mathrm{T}\left(\boldsymbol{X}_\mathrm{C}-\boldsymbol{X}_\mathrm{O}\right)=0\\\boldsymbol{n}_\mathrm{C}^\mathrm{T}\bullet \boldsymbol{n}_\mathrm{A}=\left(\boldsymbol{X}_\mathrm{C}-\boldsymbol{X}_\mathrm{O}\right)^\mathrm{T}\left(\boldsymbol{X}_\mathrm{A}-\boldsymbol{X}_\mathrm{O}\right)=0&\end{cases}.
\label{eqa8}
\end{equation}

Substitution \cref{eqa7} into \cref{eqa8}, we have the simplified form as
\begin{equation}
\begin{cases}\lambda_{\mathrm{A}}\lambda_{\mathrm{B}}\boldsymbol{x}_{\mathrm{A}}^{\mathrm{T}}\boldsymbol{\omega}\boldsymbol{x}_{\mathrm{B}}-\lambda_{\mathrm{A}}\boldsymbol{x}_{\mathrm{A}}^{\mathrm{T}}\boldsymbol{\omega}\boldsymbol{x}_{\mathrm{O}}-\lambda_{\mathrm{B}}\boldsymbol{x}_{\mathrm{B}}^{\mathrm{T}}\boldsymbol{\omega}\boldsymbol{x}_{\mathrm{O}}
+\boldsymbol{x}_{\mathrm{O}}^{\mathrm{T}}\boldsymbol{\omega}\boldsymbol{x}_{\mathrm{O}}=0\\
\lambda_{\mathrm{B}}\lambda_{\mathrm{C}}\boldsymbol{x}_{\mathrm{B}}^{\mathrm{T}}\boldsymbol{\omega}\boldsymbol{x}_{\mathrm{C}}-\lambda_{\mathrm{B}}\boldsymbol{x}_{\mathrm{B}}^{\mathrm{T}}\boldsymbol{\omega}\boldsymbol{x}_{\mathrm{O}}-\lambda_{\mathrm{C}}\boldsymbol{x}_{\mathrm{C}}^{\mathrm{T}}\boldsymbol{\omega}\boldsymbol{x}_{\mathrm{O}}
+\boldsymbol{x}_{\mathrm{O}}^{\mathrm{T}}\boldsymbol{\omega}\boldsymbol{x}_{\mathrm{O}}=0\\
\lambda_{\mathrm{C}}\lambda_{\mathrm{A}}\boldsymbol{x}_{\mathrm{C}}^{\mathrm{T}}\boldsymbol{\omega}\boldsymbol{x}_{\mathrm{A}}-\lambda_{\mathrm{C}}\boldsymbol{x}_{\mathrm{C}}^{\mathrm{T}}\boldsymbol{\omega}\boldsymbol{x}_{\mathrm{O}}-\lambda_{\mathrm{A}}\boldsymbol{x}_{\mathrm{A}}^{\mathrm{T}}\boldsymbol{\omega}\boldsymbol{x}_{\mathrm{O}}
+\boldsymbol{x}_{\mathrm{O}}^{\mathrm{T}}\boldsymbol{\omega}\boldsymbol{x}_{\mathrm{O}}=0
\end{cases},
\label{eqa9}
\end{equation}

\noindent where $\boldsymbol{\omega}=\textbf{K}^{ -\rm T}\textbf{K}^{ -1}$ is the image of the absolute conic (IAC).

\cref{eqa9} establishes a concise relationship between the IAC and the image of C2, which provides three independent constraints on the camera intrinsics. Moreover, \textbf {Proposition 1} can be easily derived as below.

\textbf {Proposition 1.} Given a calibrated camera, a C2 can be determined up to at most two different solutions from its image.

\textbf{Proof}. As an alternative parameterization to that in \cref{sec. C2para}, a C2 can also be defined by its four vertices. Given the images $\boldsymbol{x}_\mathrm{O}$, $\boldsymbol{x}_\mathrm{A}$, $\boldsymbol{x}_\mathrm{B}$ and $\boldsymbol{x}_\mathrm{C}$ of these vertices, the up-to-scale determination of this C2 is to determine a triplet of three scalars $\lambda_{\mathrm{A}}$, $\lambda_{\mathrm{B}}$ and $\lambda_{\mathrm{C}}$. Similarly, $\lambda_{\mathrm{O}}$ is set to 1. Enforcing the orthogonality constraints yields \cref{eqa9}. Since the camera intrinsics ${\mathbf K}$ is known for a calibrated camera, \cref{eqa9} changes to a ternary quadratic form. Solving this form of \cref{eqa9} leads to two solutions for the unknown triplet. 

\begin{figure}[htb]
  \centering
  \includegraphics[width=0.85\linewidth]{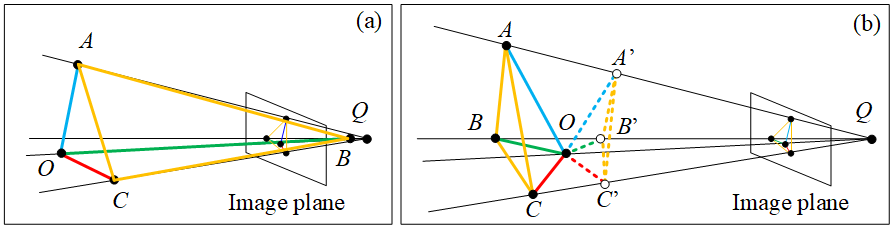}
  \caption{ Solutions of a C2 from its image: (a) configurations with only one real solution and (b) that with two real solutions, respectively, which correspond to two C2’s with different convexity.}
  \label{fig2}
\end{figure} 

There are three possible configurations for the solutions: two real different solutions, one real solution and two complex solutions, as shown in \cref{fig2}. Here we only provide a geometric interpretation about these configurations instead of rigorous proof (since it is in fact straightforward). The last two configurations correspond to a C2 located with (at least) one vertex (not the RA) close to or behind the camera center, see vertex $B$ close to the camera center $Q$ in \cref{fig2}(a), which are not possible for a real camera. Hence, we focus only on the configuration with two different real solutions, as shown in \cref{fig2}(b). These two solutions just correspond to two C2’s with the same common RA but different orientations. As shown in \cref{fig2}(b), they are “concave” and “convex” C2’s, respectively. Given another view or some prior about the C2, the “convexity” can be easily identified, see convexity check in \cref{subsec.impl}, and the only solution of C2 can thus be determined. 


\subsection{Two-view geometry}
\label{sec3_2}

\begin{figure}[hbt]
  \centering
  \includegraphics[width=0.85\linewidth]{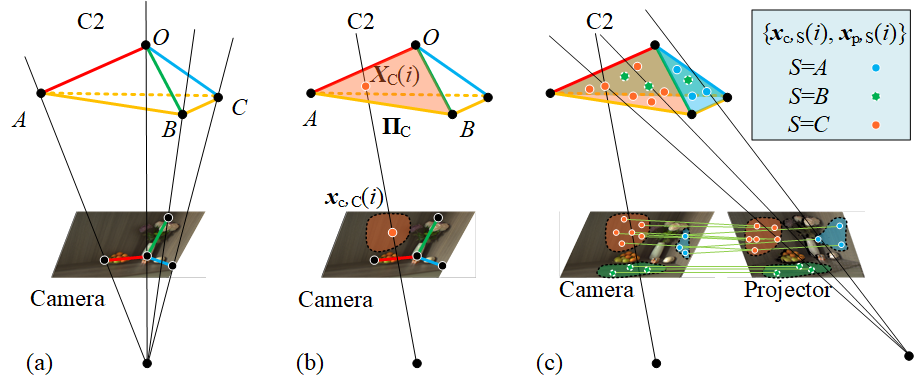}
  \caption{ Transferring from the camera to projector. (a) camera to C2, (b) C2 to $\boldsymbol{X}_{\mathrm{S}}(i)$, and (c) $\boldsymbol{X}_{\mathrm{S}}(i)$ to projector, where $S$ can be $A$, $B$ or $C$.}
  \label{fig3}
\end{figure}

Based on \textbf {Proposition 1}, the two-view geometry can be easily derived. It defines a transfer from camera intrinsics to those of the projector, as shown in \cref{fig3}. 

\textbf {Camera to C2.} Other than vertex images, matches $\{\boldsymbol{x}_{\mathrm{c,S}}(i),\boldsymbol{x}_{\mathrm{p,S}}(i)\mid i{=}1,2,$ $\ldots,N_{\mathrm{S}}\}$ on faces ${\rm{\Pi}}_{\mathrm{S}}$ of the C2 are also considered. $S$ can be $A$, $B$ or $C$, and $N_{\mathrm{S}}$ is the point number of face ${\rm{\Pi}}_{\mathrm{S}}$. Given the camera’s $\mathbf{K}_\mathrm{c}$, we first determine the unknown C2 from its vertex images according to \textbf {Proposition 1}. Those vertices $\boldsymbol{X}_\mathrm{O}$, $\boldsymbol{X}_\mathrm{A}$, $\boldsymbol{X}_\mathrm{B}$ and $\boldsymbol{X}_\mathrm{C}$ are thus computed, which are represented in the camera coordinate frame. Note that in this specific derivation the world coordinate frame coincides with the camera coordinate frame. 

\textbf {C2 to }$\boldsymbol{X}_{\mathrm{S}}(i)$. Since three points define a plane, the three faces are determined by three corresponding vertices. Accordingly, all points on there faces can be recovered by finding the intersections of the back-projecting from their image points and the respective faces. For instance, the face $\rm{\Pi }_{\mathrm{C}}$ is determined by $\boldsymbol{X}_\mathrm{O}$, $\boldsymbol{X}_\mathrm{A}$ and $\boldsymbol{X}_\mathrm{B}$. A point $\boldsymbol{X}_\mathrm{C}(i)$ satisfies both the plane equation and the back-projecting equation \cref{eqa7} as 
 \begin{equation}
 \boldsymbol{n}_{\mathrm{C}}^{\mathrm{T}}\begin{bmatrix}  \boldsymbol{X}_{\mathrm{C}}(i)\\1\end{bmatrix} =\boldsymbol{n}_{\mathrm{C}}^{\mathrm{T}}\begin{bmatrix}\lambda_{\mathrm{Ci}}\mathbf{K}_\mathrm{c}^{-1}\boldsymbol{x}_{\mathrm c,\mathrm{C}}(i)\\1\end{bmatrix}=0,
\label{eqa10}
\end{equation}
\noindent where $\boldsymbol{x}_{\mathrm{c,C}}(i)$ is the image point of $\boldsymbol{X}_\mathrm{C}(i)$ captured by the camera. $\lambda_{\mathrm{Ci}}$ is an unknown scalar. 

Solving \cref{eqa10} yields the unique $\lambda_{\mathrm{Ci}}$ and hence $\boldsymbol{X}_{\mathrm{C}}(i)$. Similarly, all points $\{\boldsymbol{X}_{\mathrm{S}}(i) \mid i{=}1,2,\ldots,N_{\mathrm{S}}\}$ on each face can be determined from their images. 

$\boldsymbol{X}_{\mathrm{S}}(i) $\textbf { to projector.} Now, these non-coplanar points $\boldsymbol{X}_{\mathrm{S}}(i) $ and their images $\boldsymbol{x}_{\mathrm{p,S}}(i)$ in the projector’s view are available. The parameter matrix $\mathbf{M}_\mathrm{p}$ for the projector can easily be solved via the Direct Linear Transformation (DLT) algorithm \cite{rc7}. The intrinsics $\mathbf{K}_\mathrm{p}$ (as well as $\mathbf{R}$ and $\boldsymbol{t}$ between the camera and projector) can obtained by decomposing $\mathbf{M}_\mathrm{p}$. 

This process establishes a transferring chain from $\mathbf{K}_\mathrm{c}$ to $\mathbf{K}_\mathrm{p}$, which can be represented by an abstract function $g(x)$ as 
\begin{equation}
\mathbf{K}_{\mathrm p}={g}\left(\mathbf{K}_{\mathrm c}\right)={g}\left(f_{\mathrm c},x_{\mathrm {c,0}},y_{\mathrm {c,0}}\right).
\label{eqa11}
\end{equation}

Additionally, if a known principal point $\boldsymbol{x}_{\rm{c,0}}=(x_{\rm{c,0}},y_{\rm{c,0}})$ of the camera is given, \cref{eqa11} further changes to a univariable function.
\begin{equation}
\mathbf{K}_{\mathrm p}={g}\left(f_{\mathrm c}\right).
\label{eqa12}
\end{equation}

According to \cref{eqa11} and \cref{eqa12}, given a C2, $\mathbf{K}_\mathrm{p}$ depends only on $\mathbf{K}_\mathrm{c}$ or $f_{\rm c}$. Accordingly, the original multi-variable (i.e. four) CPP calibration problem is simplified to a univariable estimation problem with the only unknown $f_{\rm c}$. Additionally, this relationship between $f_{\rm c}$ and $\mathbf{K}_\mathrm{2}$ is deterministic, since given a guess of $f_{\rm c}$, the corresponding $\mathbf{K}_\mathrm{p}$ can be computed uniquely without solving ambiguous polynomial equations such as Kruppa equation. This determinacy together with univariability leads to a simple yet reliable calibration algorithm.

\section{CPP self-calibration algorithm}

\subsection{Optimization objective}
According to \cref{eqa12}, we only need to determine the single unknown $f_{\rm c}$ for the CPP calibration. Please note the transferring in \cref{eqa11} from the camera to projector is invertible. Considering that the forward and backward transferring should be consistent, the camera transferring to the projector should transfer to the same camera if transferring back. This allows us to define a cycle loss between the original and transferred back camera intrinsics. Additionally, since the projector is also assumed “natural”, of which $\gamma_{\mathrm p}$ equals to 0 and $f_{\rm px}$=$f_{\rm py}$, we construct an optimization objective as 
\begin{equation}
    E\left(f_{\rm c}\right)=\underset{i=1}{\overset{7}{\sum}}E_{\rm i}\hspace{0.5cm},
\label{eqa13}
\end{equation}
where 
\begin{equation}
\begin{cases}
E_{\rm pro}=E_{\rm 1}+E_{\rm 2}=\left|\gamma_{\mathrm p}\right|+\left|f_{\rm px}-f_{\rm py}\right|\\
E_{\rm cycle}=E_{\rm 3}+E_{\rm 4}+E_{\rm 5}+E_{\rm 6}+E_{\rm 7}\\
\hspace{1cm}=\left|f_{\rm cx}^{bk}-f_{\rm cy}^{bk}\right|+\left|\gamma_{\mathrm c}^{bk}\right|+\left|f_{\rm cx}^{bk}-f_{\rm cx}^{ori}\right|+\left|f_{\rm cy}^{bk}-f_{\rm cy}^{ori}\right|+\left|\mathbf{x}_{\rm c,0}^{bk}-\mathbf{x}_{\rm c,0}^{ori}\right|
\end{cases},
\label{eqa14}
\end{equation}

\noindent where the superscripts “ori” and “bk” indicate variables from the original or back-transferred intrisics of the camera.

Solving \cref{eqa13} is a simple univariable optimization problem. Additionally, if a relatively loose feasible range for $f_{\rm c}$, say $f_{\rm c}\in[0,f_{\mathrm{max}}]$, where $f_{\rm max}$ can be set to 10000 or larger, a globally optimal solution can be estimated via an exact search \cite{rc23}. This is trivial when we discretizing the feasible range with some sampling rate $\Delta f$, say $\Delta f=1$, without sacrificing much accuracy. Alternatively, the solution can be searched via a bounded one-dimensional optimizer\cite{rc23}, which was used in our experiments due to its fast convergence and comparable stability.

\subsection{Implementation details}
\label{subsec.impl}

\textbf{Input preparation.} Our algorithm requires only matches between the camera and projector views for a C2. We firstly established correspondences using structured-light patterns \cite{r15}. These correspondences are then partitioned into three sets on faces of the C2, as shown in \cref{fig3}(c), by manually drawing respective masks, see details in \textit{Supplementary material}. 

\textbf{Inference of vertices and legs for an occluded C2.} Given only partially observed faces of a C2, the inter-view homography $\mathbf{H}_\mathrm{S}$, ${S}$= ${A}$, ${B}$ or ${C}$, induced by each face can be estimated. The leg, say $\boldsymbol{n}_{\rm A}$, between two faces, i.e. $\rm{\Pi }_{\rm B}$ and $\rm{\Pi }_{\rm C}$, can be determined using the two eigen vectors of $\mathbf{H}_\mathrm{B}*\mathbf{H}_\mathrm{C}^{-1}$\cite{rc24}. Accordingly, the $\rm RA$ can be determined from the three legs. The other three vertices can be picked manually in the legs.

\textbf{Convexity check.} In our algorithm, the convexity of a C2 is visually recognized by human or known as a prior. For instance, when reconstruction a room corner, as shown in \cref{fig1}(a), the C2 can be easily recognized as a concave one.  

\begin{figure}[htb]
  \centering
  \includegraphics[width=1\linewidth]{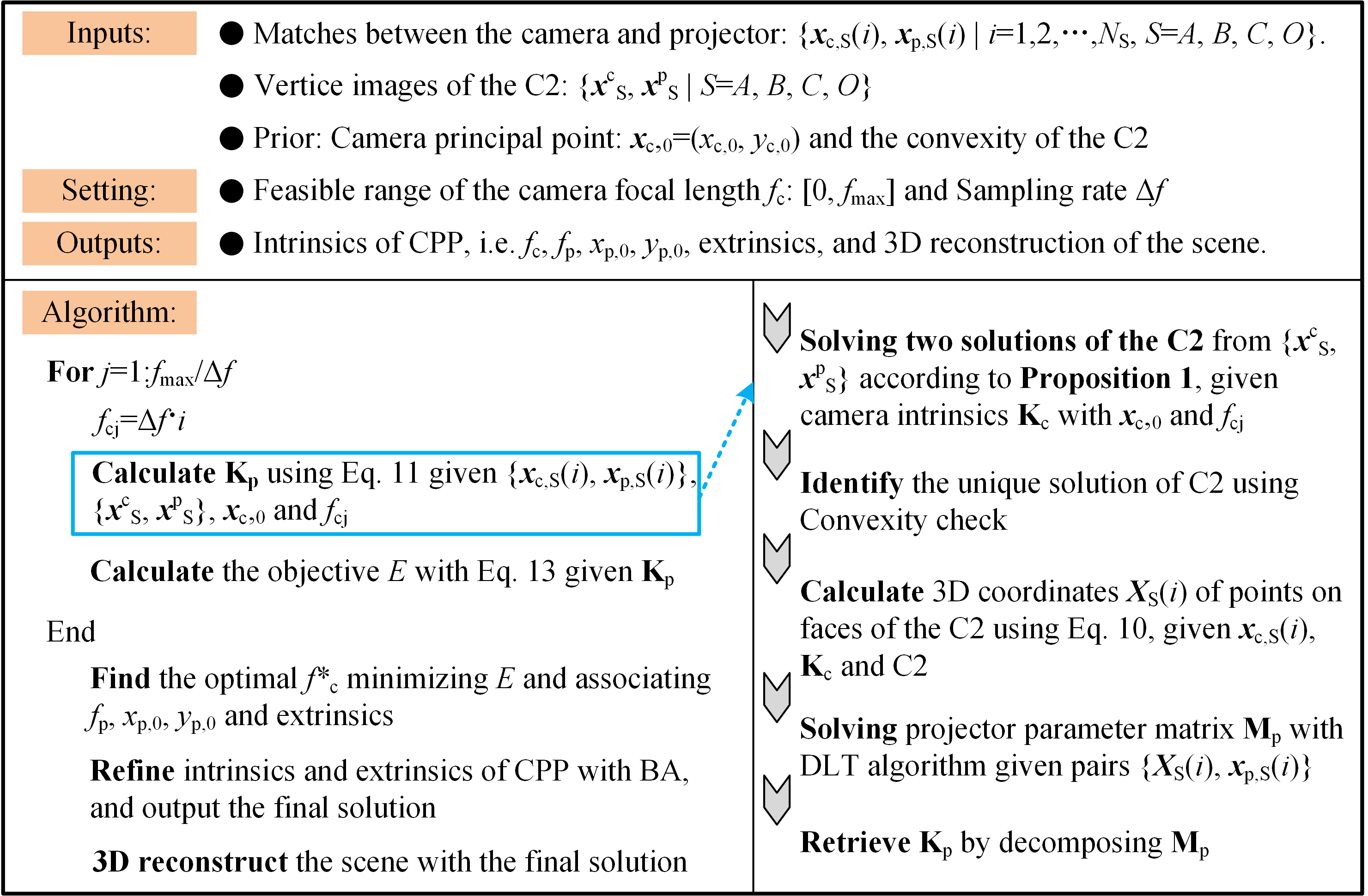}
  \caption{Pseudocode of our algorithm. Note that an exact search is used for illustration.}
  \label{fig4}
\end{figure}

The pseudocode of our algorithm is shown in \cref{fig4}.


\section{Experiments}
\label{sec:Result}

\subsection{Results on indoor scenes}

\begin{figure}[htb]
  \centering
  \includegraphics[width=0.85\linewidth]{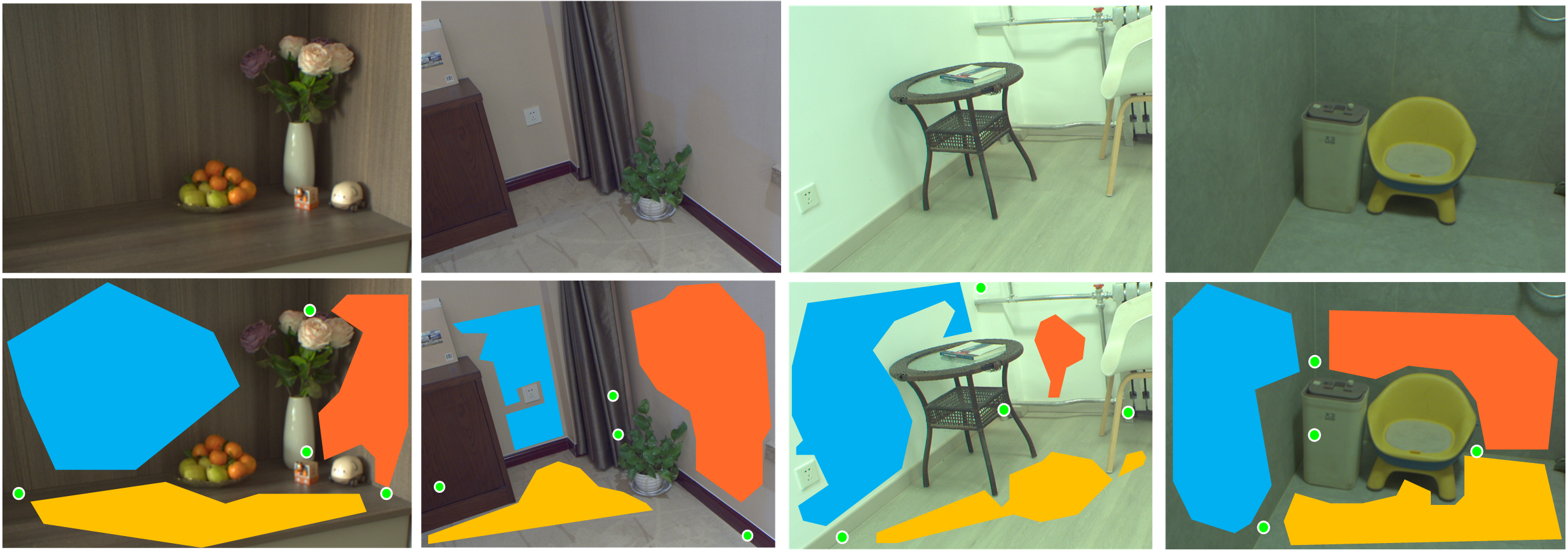}
  \caption{The camera view (the first row) and C2's with partially observed faces and their vertices of indoor scenes No. 1 - No. 4. Please note that the images are cropped to the common field of view (FOV) of the camera and projector. Readers are suggested to \textit{Supplementary material} for a full version.}
  \label{fig5}
\end{figure}

\textbf{Scenes.} Since there is no publicly available datasets of structured-light indoor images, we captured our own data using a CPP (see CCP setup below) to evaluate methods on diverse indoor scenes. As shown in the first row in \cref{fig5}, each scene contains a partially observed C2, which consists of floors and walls with repetitive or weak textures. Additionally, only limited line segments are observable in the scenes. These makes challenging scenes for CPP self-calibration and reconstruction.
To be noted, in spite of different-level occlusions, vertex images of the C2 can still be inferred from its partially observed faces, see the second row in \cref{fig5}.

\textbf{CPP setup.} A CPP with a 2448×2048 camera and a 854×480 projector was used, which contained constant but different intrinsics across scenes. The ground truths were obtained with a target-based method \cite{r20}, where the focal lengths and principal points are \{1791.1, (1256.3, 1054.3)\} and \{1247.3, (377.1, 234.0)\} for the camera and projector, respectively. More comparison results of different CPPs can be found in \textit{Supplementary material}.

\textbf{Baselines.} We compared our method with state-of-the-art methods in both two-view and multi-view configurations (see \cref{tab:1}), respectively. As baselines of traditional and learning-based self-calibration, COLMAP \cite{r2} and the recently reported DroidCalib \cite{rc21} were used. 

In the multi-view configuration, both methods estimated only camera intrinsics from about 20 views, whereas the projector was not involved since it cannot capture images. This is a typical multi-view SfM task in well-posed configuration.

In the two-view configuration, similarly, DroidCalib still only perform camera calibration but from two views, of which one is identical to the camera view used in our method and the other is close to the projector view. In contrast, COLMAP with different initial value and optimization strategies were performed for a comprehensive comparison, see \cref{tab:1}. For a fair comparison, all of them accept matches from our method as the input, i.e. $\{\boldsymbol{x}_{\mathrm{c,S}}(i),\boldsymbol{x}_{\mathrm{p,S}}(i)\mid i{=}1,2,\ldots,N_{\mathrm{S}}\}$, and then directly estimate the intrinsics of the camera and projector, respectively. The camera principal point (PP) is known a prior. In \cref{tab:1}, “C” refers to imposing the projector image center as its initial values of PP, whereas “R” means a normally random PP from the range $[1,W]\times[1,H]$. $W$ and $H$ are width and height of the projector image, respectively. COLMAP with “BA” performs additional bundle adjustment (BA) \cite{rc7} upon the result of those without “BA”. Additionally, PP's were refined in methods with “BA”, which, however, were fixed in those without “BA”.

In addition to methods above, PlaneFormer \cite{r12}, a state-of-the-art two-view 3D reconstruction method, was also compared for 3D reconstruction evaluation.

\subsubsection{Calibration result.}

As shown in \cref{tab:1}, our method achieves competitive accuracy as those of multi-view methods, reaching a mean absolute error (MAE) of 4.3\% against 0.4\% and 1.9\% for COLMAP and DroidCalib in multi-view setting. It is important to note that 20 views and only a constant camera was assumed in these baselines, which is a much simpler task than that of our method with two views of varying intrinsics. When the number of views reduces to only two, DroidCalib just failed (thus not shown in  \cref{tab:1}).  The calibration errors of two-view COLMAPs increase dramatically, especially for those of the projector intrinsics, reaching 30\% or even not converging. Additionally, since the two-view calibration is significantly ill-posed, refinement on PP with BA doesn’t improve and even reduce the accuracy, see results of COLMAP with “BA” in \cref{tab:1}. Furthermore, the performance of COLMAP demonstrates high dependency on the initial value of the projector PP. Since the image center is close to the PP, taking image center as the initial value generally leads to a more accurate calibration than those with a random value, as shown in rows marked with “C” and “R” in \cref{tab:1}. In contrast, our method direct estimate the projector PP without any initial guess, yielding more stable and accurate result.

\begin{table}[tb]
\caption{Comparison results on the calibration error.}
  \label{tab:1}
  \centering
\begin{tabular}{c|cccccccc}
\hline

\multicolumn{1}{l|}{\multirow{3}{*}{Scenes}} &
   &
  \multicolumn{2}{c}{Multi-view} &
  \multicolumn{5}{c}{Two-view} \\ \cline{3-8}
\multicolumn{1}{l|}{} &
  \multicolumn{1}{c}{Method} &
  \multicolumn{1}{c}{Colmap} &
  \multicolumn{1}{c}{DroidCalib} &
  \multicolumn{4}{c}{Colmap} &
  \multicolumn{1}{c}{Ours} \\
\multicolumn{1}{l|}{} &
  \multicolumn{1}{c}{} &
  \multicolumn{1}{c}{} &
  \multicolumn{1}{c}{} &
  \multicolumn{1}{c}{C} &
  \multicolumn{1}{c}{C-BA} &
  \multicolumn{1}{c}{R} &
  \multicolumn{1}{c}{R-BA} &
  \multicolumn{1}{c}{} \\ \hline
\multirow{4}{*}{No.1} & $f_{\rm c}$  & 0.4 & -3.1 & -2.3 & 11.5 & -70.0 &                         -60.8 & -2.4   \\
                       & $f_{\rm p}$   & - & - & 3.3 & 3.1 & -95.0 & -113.2 & 1.9 \\
                       & $x_{\rm p,0}$ & - & - & -13.2 & -12.5 & 22.6 & 38.6 &  -9.3 \\
                       & $y_{\rm p,0}$ & - & - & -2.6 & -1.7 & -67.4 & -116.8 & -10.2 \\ \hline
\multirow{4}{*}{No.2} & $f_{\rm c}$   & 0.4 & -1.4 & -68.5 & -33.0 & -47.4 & -56.5 & -4.6 \\
                       & $f_{\rm p}$   & - & - & -382.1 & -497.5 & -966.6 & -1634.8 & 0.1 \\
                       & $x_{\rm p,0}$ & - & - & -13.2 & 72.3 & 22.6 & -46.1 & -9.1 \\ 
                       & $y_{\rm p,0}$ & - & - & -2.6 & -53.2 & -67.4 & -108.5 & 4.2 \\
                       \hline
\multirow{4}{*}{No.3}   & $f_{\rm c}$   & -0.1 & -1.6 & -41.3 & -32.6 & -43.9 & -39.3 & -3.0 \\
                       & $f_{\rm p}$   & - & - & -128.7 & -99.5 & -164.3 & -143.2 & -0.8\\
                       & $x_{\rm p,0}$ & - & - & -13.2 & -6.1 & 22.6 & 26.7 & -7.3 \\
                       & $y_{\rm p,0}$ & - & - & -2.6 & -18.4 & -67.4 & -65.6 & -4.9 \\ \hline
\multirow{4}{*}{No.4} & $f_{\rm c}$   & 0.6 & -1.6 & -37.2 & -26.7 & -103.4 & -37.2 & 1.3 \\
                       & $f_{\rm p}$   & - & - & -68.9 & -64.9 & -181.3 & -178.3 & 6.6 \\
                       & $x_{\rm p,0}$ & - & - & -13.2 & -18.6 & 22.6 & 16.1 & -3.4\\
                       & $y_{\rm p,0}$ & - & - & -2.6 & 32.8 & -67.4 & -148.9 & 0.1 \\ \hline

\multicolumn{1}{l|}{} &  MAE    & \textbf{0.4}  & 1.9  & 49.7  &  61.5 & 127.0 &  176.9& \textbf{4.3} \\ 

\hline
\end{tabular}
\end{table}

\begin{figure}[htb]
  \centering
  \includegraphics[width=1\linewidth]{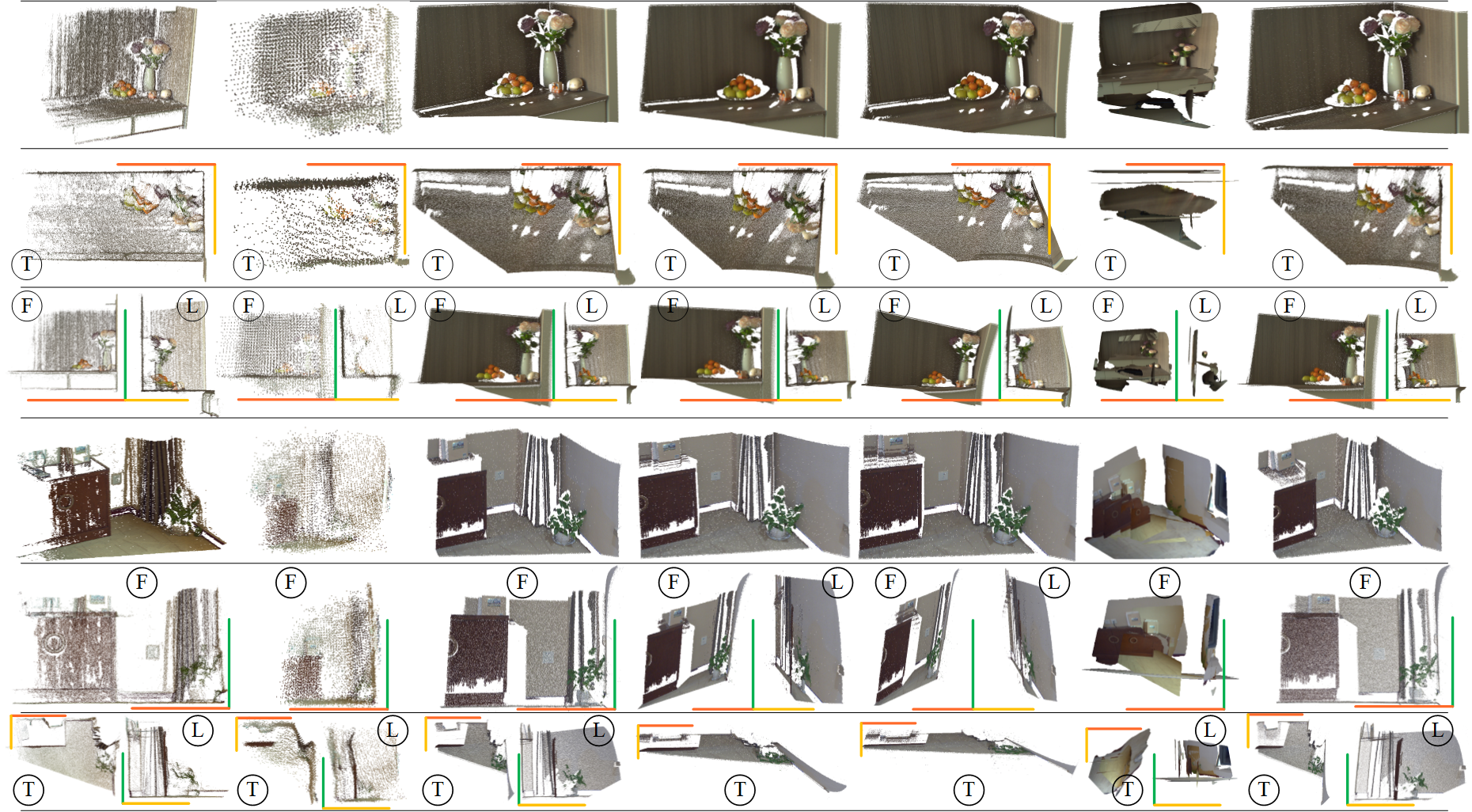}
  \caption{Reconstruction of two scenes. (\textit{From left to right}) Multi-view COLMAP, DriodCalib, two-view COLMAP with GT CPP parameters, C-BA, R-BA, PlaneFormer and ours. Each row showcases the result from the top, front and left views with markers “T”, “F” and “L”, respectively. Three orthogonal axes are shown in different colors.}
  \label{fig6}
\end{figure}

\begin{figure}[hb]
  \centering
  \includegraphics[width=1\linewidth]{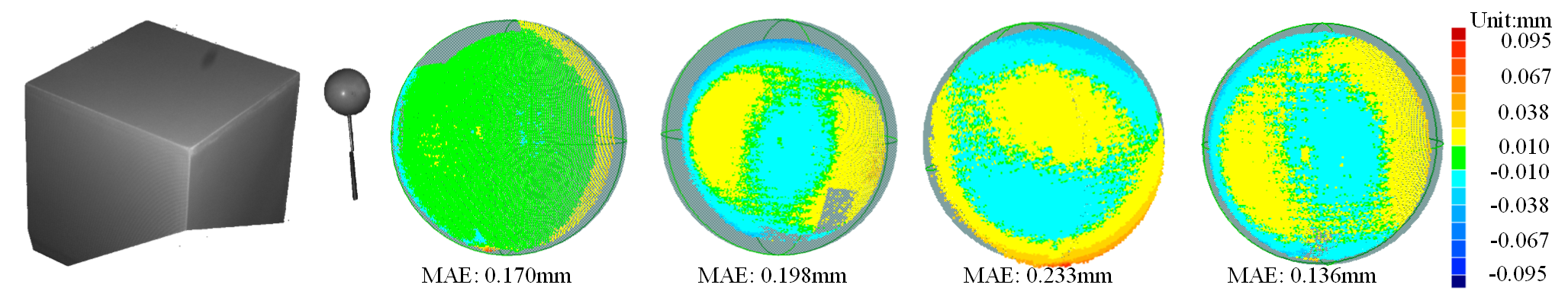}
  \caption{(\textit{From left to right}) The reconstructed cuboid corner and sphere, heatmaps of COLMAP with GT CPP intrinsics, C-BA, R-BA and ours, respectively.}
  \label{fig7}
\end{figure}

\subsubsection{Reconstruction result.}

As expected, multi-view COLMAP achieves the most accurate result, whereas those from its learning-based counterpart are “noisier” with noticeable distortions on orthogonal planes, such as the floor and wall in the second scene in \cref{fig6}. In contrast to both methods, our method produced reconstruction with similar fidelity but higher density from two views. 
Additionally, compared with two-view methods such COLMAPs with C-BA, R-BA or PlaneFormer, which generate significantly distorted reconstructions due to inaccurate calibration, as shown in the fourth and fifth columns in \cref{fig6}, or tend to fail due to mis-detection and matching of planes, as shown in the sixth column in \cref{fig6}, our method demonstrated noticeable superiority over baselines. Furthermore, comparable reconstruction accuracy was achieved against COLMAP with GT CPP parameters.

Since the orthogonality of faces of the C2 has been exploited inherently in our method, for a fairer comparison, we further quantitatively compared the reconstruction accuracy of a high-precision sphere. As shown in \cref{fig7}, the corner of a cuboid was used for calibration, and a 50.8-mm sphere was used for evaluation. Please note that the sphere has never been used for calibration. We use MAE and heatmaps of sphere-fitting errors to access the reconstruction accuracy. While the heatmap demonstrates the COLMAP with GT achieves balanced performance with more even error distribution across the sphere surface, our method reaches the lowest MAE among them, demonstrating a competitive performance.

\begin{figure}[htbp]
  \centering
  \includegraphics[width=1\linewidth]{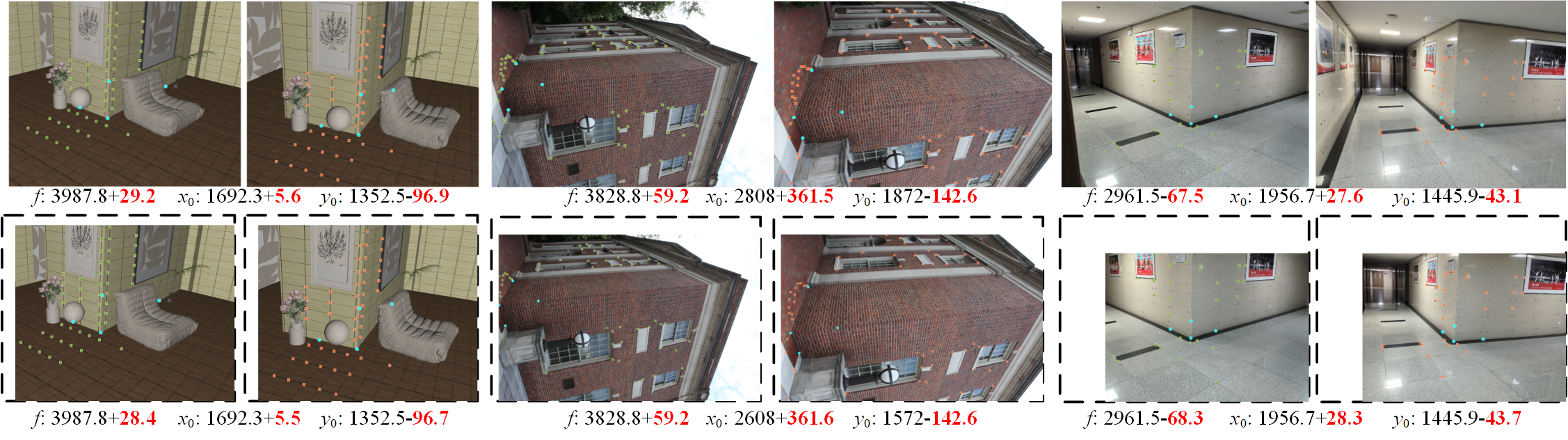}
  \caption{(\textit{From left to right}) Two views of a simulated indoor scene and two real scenes (the middle one is from the “graham” dataset in \cite{r2}). The first row shows the original images with the principal point (PP) on the image center, which are cropped to produce PP shift, as shown in the second row, where dashed lines indicate the original image boundary. The unit for all values (including calibration error in red) is pixel.}
  \label{fig8}
\end{figure}

\subsection{Extension to two-view SfM}

Instead of limited in indoor scenes with a CPP, our method can be easily adapted to more general scenarios with only cameras, where we try to self-calibrate cameras in a two-view SfM problem. A constant but unknown camera across views was assumed. With simple modification on the objective in \cref{eqa13}, see details in \textit{Supplementary material}, the ill-posed problem is solved, as shown in the first row in \cref{fig8}. Furthermore, since additional constraints can be derived from the assumption of a constant camera, no prior about the camera principal point is required. This allows to estimate all camera intrinsics even from a pair of cropped images, as shown in the second row in \cref{fig8}. Surprisingly, our method achieved consistent accuracy on cropped images as the original ones, which demonstrates considerable stability and robustness to image cropping. It provides potential solution for reliable camera self-calibration in two-view SfM.

\subsection{Impact of the optimization objective and density of matches}

\begin{figure}[htbp]
  \centering
  \includegraphics[width=1\linewidth]{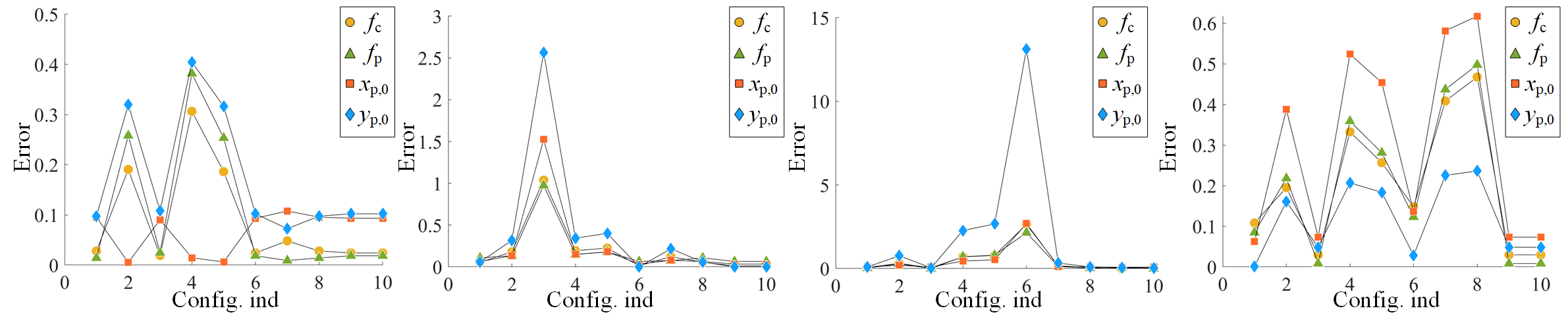}
  \caption{Results with different configs. of optimization objective for four scenes.}
  \label{fig9}
\end{figure}

We further investigated the impact of different configurations of the optimization objective, to determine which one is more significant for the calibration. Ten configurations were evaluated, of which the first seven corresponded to $E_{\rm i}, i=1,2,…,7$, and the last three were $E_{\rm pro}$, $E_{\rm cycle}$ and $E$ in \cref{eqa13} and \cref{eqa14}, respectively. As shown in \cref{fig9} and \cref{tab:2}, the last two configurations contribute dominantly to the calibration, which achieved the lowest errors across scenes. This indicates that the cycle loss is much more significant than $E_{\rm pro}$. The sum of $E_{\rm pro}$ and $E_{\rm cycle}$, i.e. $E$, achieved comprehensively best performance, reaching the top calibration accuracy.

\begin{table}[htbp]
\caption{MAE of different configurations}
    \label{tab:2}
    \centering
    \setlength{\tabcolsep}{2.5mm}{
\begin{tabular}{ccccccccccc}
\hline
 {Config} & 1 & 2 & 3 & 4 & 5 & 6 & 7 & 8 & 9 & 10 \\
\hline
 {Error/\%} & 6.7 & 25.4 & 41.8 & 46.9 & 48.2 & 133.8 & 19.6 & 16.5 & 4.4 & 4.3 \\
\hline
\end{tabular}}
\end{table}

We also investigated the impact of the number of matches. The original matches were downsampled by different rates, where the average number for the four scenes were reduced from an order of magnitude 9×$10^{4}$ to that of only 100. As shown in \cref{tab:3}, our method performs consistently well with different downsampling rates, demonstrating robustness to the number of matches.

\begin{table}[!h]
\caption{MAE of different downsampling rates}
    \label{tab:3}
    \centering
    \setlength{\tabcolsep}{2.9mm}{
\begin{tabular}{ccccccccc}
\hline
 {Downsampl.rate} & 1 & 10 & 20 & 100 & 200 & 300 & 500 & 100 \\
\hline
 {Error/\%} & 4.29 & 4.32 & 4.34 & 4.29 & 4.23 & 4.33 & 4.31 & 4.22 \\
\hline
\end{tabular}}
\end{table}

\subsection{Degenerated configuration and limitation}
There are still some limitations for the proposed method. On the one hand, our method fails in some degenerated configuration, for instance, when at least one of the faces of a C2 passes through the camera center. This is just the case where all points on a face are imaged to the same line. On the other hand, our method requires accurate matches of three segmented faces of a C2. To achieve this, we manually segment images and use structured light patterns to establish reliable and accurate correspondence across views. In the future work, we will further develop algorithms for automatically detecting faces of a C2 and their matches, to fully automate the calibration and reconstruction.

\section{Conclusion}
This paper proposes to use the C2, a common and simple structure in most daily indoor scenes, to solve the ill-posed two-view CPP self-calibration problem with varying intrinsics across views. The view geometry of a C2 is derived, from which sufficient constraints can be constructed. These constraints allow to simplify the complex multi-variable estimation problem of CPP calibration to a much simpler uni-variable searching one, resulting a reliable and accurate calibration and thus enabling indoor 3D reconstruction with an unknown CPP. Compared with both traditional and learning-based state-of-the-art methods, the proposed method has demonstrated significant improvement on both calibration and reconstruction accuracy. Additionally, the proposed method also demonstrates promising potential for similar tasks such as camera self-calibration in sparse-view SfM. 

\par\vfill\par

%
%
\bibliographystyle{splncs04}
\bibliography{references}
\end{document}